\documentclass[10pt,twoside]{article}

\usepackage{times}
\usepackage[utf8]{inputenc}
\usepackage[T1]{fontenc}
\usepackage{graphicx}

\usepackage{siunitx}
\usepackage{hyperref}
\usepackage{cleveref}
\usepackage{coria-taln2025}
\usepackage[french]{babel} %frenchb deprecated, use french instead

\title{Structuration Automatique de la Posologie en Français : Quel rôle pour les LLMs ?}

\author{
  \vspace{-0.5cm}
  Natalia~Bobkova\up{1} \quad Laura~Zanella-Calzada\up{1} \quad Anyes~Tafoughalt\up{1,2} \quad Raphaël~Teboul\up{1} \quad François~Plesse\up{1}\footnote{Was at Posos at the time of his contribution.} \quad Félix~Gaschi\up{1}\\
  {\small
    (1) SAS Posos, (2) Sorbonne Université, Paris, France \\
    \texttt{felix@posos.co}
}} 

\begin{document}
\maketitle

\vspace{-0.5cm}

\resume{ La structuration automatique de posologie est essentielle pour
  fiabiliser la médication et permettre une assistance à la prescription
  médicale. Les textes de prescriptions en français présentent très souvent des
  ambiguïtés, des variabilités syntaxiques, et des expressions colloquiales, ce
  qui limite l'efficacité des approches classiques de machine learning. Nous
  étudions ici l'emploi de Grands Modèles de Langages (\textit{LLM}) pour structurer les
  textes de posologie en comparant des méthodes fondées sur le
  \textit{prompt-engineering} et le \textit{fine-tuning} de LLM avec un
  système "pré-LLM" fondé sur un algorithme de reconnaissance et liaison
  d'entités nommées (NERL). Nos résultats montrent que seuls les LLM fine-tunés
  atteignent la précision du modèle de référence. L'analyse des erreurs révèle
  une complémentarité des deux approches : notre NERL permet une structuration
  plus précise, mais les LLMs captent plus efficacement les nuances sémantiques.
  Ainsi, nous proposons le modèle hybride suivant : faire appel à un LLM en cas
  de faible confiance en la sortie du NERL (<0.8) selon notre propre score de
  confiance. Cette stratégie nous permet d'atteindre une précision de 91\% tout
  en minimisant le temps de latence. Nos résultats suggèrent que cette approche
  hybride améliore la précision de la structuration de posologie tout en
  limitant le coût computationnel, ce qui en fait une solution scalable pour une
  application clinique en conditions réelles. }

\abstract{Automatic Posology Structuration: What role for LLMs?}{ \vspace{-0.2cm}Automatically
  structuring posology instructions is essential for improving medication safety
  and enabling clinical decision support. In French prescriptions, these
  instructions are often ambiguous, irregular, or colloquial, limiting the
  effectiveness of classic ML pipelines. We explore the use of Large Language
  Models (LLMs) to convert free-text posologies into structured formats,
  comparing prompt-based methods and fine-tuning against a "pre-LLM" system
  based on Named Entity Recognition and Linking (NERL). Our results show that
  while prompting improves performance, only fine-tuned LLMs match the accuracy
  of the baseline. Through error analysis, we observe complementary strengths:
  NERL offers structural precision, while LLMs better handle semantic nuances.
  Based on this, we propose a hybrid pipeline that routes low-confidence cases
  from NERL (<0.8) to the LLM, selecting outputs based on confidence scores.
  This strategy achieves 91\% structuration accuracy while minimizing latency
  and compute. Our results show that this hybrid approach improves structuration
  accuracy while limiting computational cost, offering a scalable solution for
  real-world clinical use. }

\motsClefs
  {TAL, Larges Modèles de Langage, TAL clinique, TAL médical} 
  {NLP, Large Language Models, clinical NLP, medical NLP} 

\newcommand{\dataset}{\textsc{MedPosoSF}~}
\newcommand{\datasetFullname}{\textbf{Med}ical \textbf{Poso}logy \textbf{S}tructuration in \textbf{F}rench~}

%%%% IMPORTANT ! %%%%
%\acceptedArticle[accepted]{Nom de la Conférence / Revue} % Pour une contribution acceptée pour publication
%\acceptedArticle[submitted]{Nom de la Conférence / Revue} % Pour une contribution soumise
\acceptedArticle[submitted]{MLP-LLM à TALN 2025} % Pour une contribution originelle

\section{Introduction}
Structuring posology instructions is essential for enhancing medication safety and enabling clinical applications, including decision support systems \cite{Elhaddad2024-ya}. In French prescriptions, posologies are frequently expressed in diverse, ambiguous, and sometimes colloquial language, which complicates their automatic structuration.

Traditional information extraction pipelines, typically based on Named Entity Recognition (NER) and rule-based post-processing, offer reliable performance but struggle with linguistic variability and unexpected formulations \cite{Liang2019-mz,mednerf}. In contrast, Large Language Models (LLMs) have recently shown strong potential for medical text understanding \cite{hu2023gptner,biana2024vaner}, though their application to specialized tasks like posology structuration remains largely unexplored.

In this work, we investigate the use of LLMs to structure French posology instructions into standardized formats. We benchmark several open-source and proprietary LLMs using prompt engineering techniques such as chain-of-thought, few-shot, and contrastive prompting. To further improve performance, we apply lightweight fine-tuning on synthetic posology data.

Our results show that while prompt engineering narrows the gap, fine-tuning is necessary for LLMs to match the performance of our robust internal baseline (based on Named Entity Recognition and Linking, or NERL). Through detailed error analysis, we highlight complementary strengths between rule-based systems and LLMs. Finally, we propose a hybrid strategy that leverages model confidence scores to combine their outputs, offering a practical and efficient solution for real-world deployment.

We summarize our contributions as follows:
\begin{enumerate}
    \item We present a novel dataset, \dataset, for evaluating LLM-based structuration of French posology instructions, extending beyond standard NER by requiring full structured outputs.
    \item We benchmark a range of LLMs, open and proprietary, using prompt engineering and fine-tuning, and analyze their ability to handle the linguistic variability of real-world prescriptions.
    \item We propose a confidence-based hybrid pipeline that combines a rule-based NERL system with LLMs, achieving improved accuracy (91\%) while controlling latency and resource usage.
\end{enumerate}

\section{Related Work}
%we need to cite work using LLMs for information extraction in the medical
%domain, and insist on the fact that they do it for simpler tasks (namely NER,
%and probably some classification tasks). The whole related work part should end
%with a paragraph explaining that we are the first to use LLMs for the
%structuration of posology instructions in French.
Research on structuring clinical texts often begins with simpler tasks such as NER, which serves as a foundational step for more complex transformations, including the structuring of dosage instructions.

\paragraph{Named Entity Recognition in the Medical Domain}
NER is a core task in biomedical Natural Language Processing (NLP), aiming to identify entities such as medications, dosages, symptoms, or diseases within clinical texts. Over time, several approaches have been developed, each attempting to overcome the limitations of the previous ones.

Early methods were mainly based on handcrafted rules and domain-specific dictionaries. These systems typically relied on keyword selection and pattern-matching techniques to extract relevant terms from text ~\cite{8200102}.
While they offered high precision in specific contexts, their recall was often limited, and they lacked adaptability. Moreover, maintaining and updating these resources required significant effort and deep domain expertise ~\cite{ijerph17082687}, making them difficult to scale across different datasets or languages.

To address these issues, statistical machine learning models like Hidden Markov Models (HMMs) \cite{shen-etal-2003-effective} and Conditional Random Fields (CRFs) emerged ~\cite{he-wang-2008-chinese, settles-2004-biomedical, 10.1136/amiajnl-2011-000163}.
CRFs, in particular, became widely used for sequence labeling tasks, as they modeled label dependencies and improved prediction consistency. In the French context, \citet{bigeard2015automatic} proposed a hybrid system combining CRFs and handcrafted rules to extract numerical values from unstructured EHRs. Their work highlighted challenges specific to the French language, such as ambiguous units and linguistic variability, emphasizing the need for language-specific approaches in biomedical NLP.

However, they still depended heavily on feature engineering and struggled to capture long-distance relationships between tokens.
A major shift came with the introduction of deep learning. Long Short-Term Memory (LSTM) networks allowed models to retain information across longer sequences ~\cite{peng2017improvingnamedentityrecognition, Li2016RecognizingBN}.
Their bidirectional versions, BiLSTMs, further improved context modeling by processing text in both forward and backward directions ~\cite{wang2015unifiedtaggingsolutionbidirectional}.
The combination of BiLSTM with CRFs led to the well-known BiLSTM-CRF architecture, which merged contextual encoding with structured prediction ~\cite{huang2015bidirectionallstmcrfmodelssequence, lample-etal-2016-neural}. This hybrid approach became a reference in the field, especially when combined with domain-adapted embeddings such as those trained on PubMed (e.g., Word2Vec or GloVe). However, the use of static embeddings still limited the model’s ability to adapt to varying contexts.

More recently, the use of self-attention mechanisms ~\cite{vaswani2023attentionneed} and transformer-based architectures has significantly improved performance in NER. Unlike RNNs, self-attention can model global dependencies in a single pass, making it especially well-suited for capturing the structure of clinical narratives.

In this context, transformer-based models like BERT ~\cite{devlin2019bertpretrainingdeepbidirectional} and its biomedical variants have set new standards for medical NER. BioBERT \cite{lee2020biobert}, pretrained on large biomedical corpora, ClinicalBERT \cite{alsentzer2019publicly}, fine-tuned on clinical notes (MIMIC-III), and SciSpacy \cite{neumann2019scispacy}, built on PubMed articles, have all achieved state-of-the-art results. These models benefit from contextual embeddings tailored to biomedical language, drastically reducing the need for manual feature design and improving robustness in handling ambiguous or complex terms.

\paragraph{Using Large Language Models (LLMs) for Medical NER}
With the rise of general-purpose Large Language Models (LLMs) like GPT-3.5 and
GPT-4, researchers have started exploring their use for biomedical NER.
~\citet{hu2023gptner} evaluated GPT-3.5 and GPT-4 on two clinical NER tasks:
identifying medical problems, tests, and treatments from clinical notes, and
extracting adverse events from safety reports. Under
this setup, GPT-4 performance is close to those of domain-specific models like
BioClinicalBERT.

To improve performance, some researchers have proposed adapting LLMs more explicitly to NER. For instance, ~\citet{wang2023gptner} introduced GPT-NER, a framework that reframes NER as a text generation task. This allows the model to directly output labeled sequences, rather than relying on token classification. Their results suggest that this generation-based formulation can be especially effective in low-resource scenarios, where annotated data is scarce.

More recently, \citet{biana2024vaner} presented VANER, a system built on top of LLaMA2. It combines domain-specific instructions and external biomedical knowledge to improve entity recognition. VANER outperformed previous LLM-based methods and even surpassed some traditional BioNER baselines on multiple datasets.

In parallel, \citet{naguib-etal-2024-shot} conducted a large-scale evaluation of generative and masked LLMs for few-shot clinical NER across English, French, and Spanish. Their results show that while prompt-based models perform well on general NER, they are outperformed in the clinical domain by lighter, fine-tuned models like BiLSTM-CRF.

Overall, while general-purpose LLMs may not yet consistently outperform specialized models like BioClinicalBERT, studies show that prompt engineering and hybrid strategies (e.g., adding domain knowledge) can significantly boost their effectiveness for clinical NER.

\paragraph{Structuring Posology Instructions: Beyond NER}
While NER focuses on identifying relevant entities in text, dosage structuring advances further by transforming 
complex dosage instructions into structured formats (e.g., JSON), facilitating integration into clinical information systems.

A key milestone in the structuration of medication-related data was the i2b2 Medication Challenge ~\cite{10.1136/jamia.2010.003947}, which defined subtasks such as drug name, dosage, frequency, and route extraction. It remains a standard benchmark for evaluating medication information extraction systems, particularly in English clinical texts.

So far, only a few studies have looked into applying LLMs to tackle related tasks. For instance, \citet{van2024rx} 
introduced Rx Strategist, a multi-agent LLM pipeline augmented with knowledge graphs and search strategies to verify 
prescriptions by analyzing indications, dosages, and potential drug interactions. This approach achieved performance 
comparable to that of experienced pharmacists.
Similarly, \citet{isaradech2024zero} investigated zero and few-shot NER and text expansion in medication prescriptions 
using ChatGPT 3.5 and highlighting 
the potential of LLMs in handling varied and ambiguous language in prescriptions.

\citet{silva2025adaptive} developed an adaptive LLM-based intelligent medication assistant aimed at supporting 
decision making in antidepressant prescriptions, demonstrating the utility of fine-tuned language models in 
clinical decision support workflows and emphasizing the importance of structured dosage information for 
clinical decision making. \citet{haaker2025approaches} conducted a comparative evaluation of five approaches 
for extracting daily dosages, including parsing techniques, LLMs (GPT-4o), and the rule-based system RxSig, 
reporting slightly better performance with GPT-4o than with RxSig, along with notable differences in sensitivity and 
computational demands.

These studies highlight the growing role of LLMs in transforming how posology instructions are interpreted and structured, 
offering richer and more flexible representations than traditional NER approaches.

Our work extends this research approach to French prescriptions, marking, to our knowledge, one of the first efforts to apply 
LLMs to the structured extraction of posology instructions.
\section{Methodology}

\subsection{The posology structuration task}

We release \datasetFullname~(\dataset), a posology structuration test dataset
for converting free-text posology instructions into structured JSONs, through
which we normalize elements of the posology instruction like the frequency, the
intake dose, or the duration of the treatment.

We collected sentences containing dosage instructions from a private set of
French scanned typewritten prescriptions. Following \citet{mednerf}, those
prescriptions were passed through an Optical Character Recognition (OCR)
system\footnote{\url{https://cloud.google.com/vision}}. Only the sentences
containing posology instructions and no other patient information were kept,
ensuring anonymization of the data\footnote{To ensure anonymization, we manually checked each sentence in the final dataset and removed any that contained personal info (name, social security number, doctor ID)}.

Those sentences were then manually annotated by medical experts. Contrary to
simpler structuration tasks like NER, the task is not a classification, but is
more akin to a text-to-JSON conversion. Thus, instead of tagging spans of text
with relevant labels, annotators were asked to fill a form for each relevant
spans of text (e.g. "1 ampoule tous les 2 mois pendant 4 mois"), leading to a
JSON with the following important fields (The complete schema is shown in
Appendix \ref{sec:schema})\footnote{Annotations were performed by two annotators
using an internally-developped tool that converts the output of an user-friendly
form into JSON.}:

\begin{itemize}
\item \texttt{quantity\_and\_rate}: dict, contains the amount of intake units to take at once
\begin{itemize}
  \item \texttt{value}: float, the amount of unit to take at once (1.0 in our example)
  \item \texttt{unit}: str (optional), the type of unit ("ampoule(s)")
  \item \texttt{code}: str (optional), the SNOMED code for the unit ("413516001")
\end{itemize}
\item \texttt{timing}: dict, contains the timing with which each intake must take place
\begin{itemize}
\item \texttt{frequency}: int, an integer for the number of intake for the given period (1)
\item \texttt{period}: int, the number of period\_unit to observe before repeating the intake (2)
\item \texttt{period\_unit}: str, the time unit used for measuring the period between intakes ("month")
\end{itemize}
\end{itemize}

Our evaluation metric is a query-wise exact accuracy, which measures the
proportion of queries for which each field is correctly
predicted\footnote{SNOMED codes were not considered in the evaluation, although
some LLMs are sometimes able to find the right code, we leave this named entity linking
step for future work.}. We publicly release our manually annotated dataset
and code for evaluating the predictions of any
model\footnote{\url{https://github.com/posos-tech/posology-structuration-task}}.

\subsection{Baseline: an internal pre-LLM system}

We design our baseline as a hybrid system that combines transformer-based and
rule-based NER with embedding-based Named Entity Linking (NEL). The NER module
consists of a fine-tuned \texttt{microsoft/mdeberta-v3-base}
\cite{he2023debertav3improvingdebertausing} model. It is trained on an internal
manually annotated dataset\footnote{For the annotation process, annotators used
an internally developed tool that allows them to select relevant textual
information and assign the appropriate tags.}, which we unfortunately cannot
release due to issues of confidentiality. This dataset contains 1,699 sentences
annotated with 5,845 entities (Full list of entities in Appendix
\ref{sec:ner-training}). 

Some entity types are rare, for example, there are only 19 entities of type
\texttt{TIME\_OF\_DAY} (like "à 08h") in the entire training set. This requires
some form of data augmentation. We thus define a set of rules to generate
synthetic training examples that include those rarer entities.

For some of the entities detected with the NER, we need to link them to relevant
concepts using NEL. We link recognized dose units to normalized concepts from the
SNOMED CT terminology. Our linking pipeline embeds
candidate concepts using \texttt{FastText} \cite{joulin2016fasttext} word
vectors aggregated into sentence-level representations. We retrieve the closest concepts based on
cosine similarity. 

However, extracting and linking entities is not sufficient to produce an
arborescent JSON describing the posology. An internal rule-based post-processing
system allows to combine the detected entities into a structured JSON. We choose
not to disclose our entire set of rules, as they were developped internally and
iteratively across several years. A typical example of such rules is that, in "3
cp le matin" two entities are initially detected: a dose (3 cp) and an entity of
type \texttt{WHEN} (le matin). The rule-based system will then combine them into
a single JSON object with the \texttt{quantity\_and\_rate} filled with the
information from the dose and the \texttt{frequency} (1), the \texttt{period}
(1) and the \texttt{period\_unit} (day) are inferred from the \texttt{WHEN}
entity.

\subsection{Confidence score}

Following the method proposed by~\citet{9679138}, we crafted a confidence score aimed to assess risk, defined by three kinds of uncertainties:
model, aleatoric and epistemic. Considering the structuration model, i.e. the NERL baseline, as a black-box classification model,
we trained an ensemble of Gradient Boosted classifiers as a meta-learner to estimate these uncertainties. Each model is a regression tree that outputs a score for the correct or incorrect label.
The final prediction is obtained by averaging the outputs of all models in the ensemble. The different uncertainties were evaluated as follow: 
\begin{itemize}
  \item \textbf{Model uncertainty}: out of NERL model information, estimated using the final model score.
  \item \textbf{Aleatoric uncertainty}: variability of data points, and noise caused in our case by mistakes of OCR for example, evaluated by averaging the entropy of each smaller model.
  \item \textbf{Epistemic uncertainty}: systematic gaps in the training samples, it mostly arises when outliers are present within the training dataset. It was deduced from the total uncertainty, estimated from the entropy of the model outputs.
\end{itemize}

We trained the confidence score on two datasets with a total of 600 sentences:
each sentence is paired with the JSON output produced by our NERL baseline and
manually annotated as "correct" or "incorrect" according to the ability of our
model to correctly structure them. The confidence score model is then trained,
using the sentence and the JSON output as features, to predict the correctness
of the output.

\subsection{LLM approaches}

While the aforementionned pre-LLM system is effective for structuring posology
instructions (see results), it is limited by its reliance on handcrafted rules,
the need for extensive domain knowledge, and requires a training dataset for the
NER. To address these limitations, we explore the use of Large Language Models
(LLMs) to automate the structuration process, without using any handcrafted
rules or training dataset other than a synthetic one.

\paragraph{Prompt engineering}\mbox{}\

We explore the integration of Large Language Models (LLMs) to improve the pososlogy structuration task, focusing on prompt engineering as an initial approach. Our experiments involve several open-source models, such as LLaMa 3.2~\cite{grattafiori2024llama}, Gemma 2~\cite{team2024gemma}, Mistral 7B~\cite{albert2023mistral}, and Phi 3.5~\cite{abdin2024phi}; as well as proprietary models, particularly the Gemini family~\cite{team2024gemini}, deployed via Vertex AI.

To optimize model behavior, we employ several prompting strategies: chain-of-thought prompting~\cite{wei2022chain} which encourages a step-by-step reasoning; few-shot prompting~\cite{brown2020language} to embed structured examples in the prompt; reformulation instructions~\cite{deng2023rephrase} which direct the model to clarify or rephrase the posology instruction before structuring it; contrastive prompting~\cite{chia2023contrastive} which includes examples of common errors to help the model avoid frequent pitfalls. The prompts used are described in detail in Appendix~\ref{sec:prompt-details}.

\paragraph{Fine-tuning with synthetic data}\mbox{}\

To further improve LLM performance, we apply lightweight fine-tuning techniques such as adapter-based methods~\cite{houlsby2019parameter}. Given the scarcity of annotated prescription data, we construct a synthetic dataset of over 1,600 posologies, structured using our NERL baseline, alleviating the need for any manual annotation\footnote{Although the baseline itself required annotated data to function}.

From this set, approximately 1,200 are filtered out based on cofidence score (>0.7). Despite being synthetic, the dataset reflects the structural and linguistic diversity of real-world posologies and serves as an effective foundation for task-specific model tuning~\cite{gunasekar2023textbooks}.

\section{Experiments and Results}

\subsection{Models comparison}\mbox{}\

Our first objective is to select a suitable LLM for the posology structuration task. We run initial baseline tests using simple prompts across open-source (LLaMA 3.2\footnote{3.21B parameters, Q4\_K\_M quatisation}, Gemma 2\footnote{9.24B paramerters, Q4\_0 quatisation}, Mistral 7B\footnote{7.25B parameters, Q4\_0 quatisation}, Phi 3.5\footnote{3.82B parameters, Q4\_0 quatisation}) and closed-source (Gemini family) models, evaluating both their accuracy and inference latency\footnote{Except for Gemini models which run on Google servers, all tests for Ollama models were performed on the same GPU} on manually curated data set (Table~\ref{tab:llm-eval}). 

\begin{table}[ht]
  \centering
  \begin{tabular}{lrr}
  \hline
  \textbf{Model} & \textbf{Average Accuracy} & \textbf{Average Latency} \\
  \hline
  Vertex AI - Gemini 1.0 Pro   & 38\%  & 14 min \\
  Vertex AI - Gemini 1.5 Pro   & 47\%  & 11 min \\
  Vertex AI - Gemini 1.5 Flash & 42\%  & 3 min  \\
  Ollama - Gemma 2             & 51\%  & 20 min \\
  Ollama - Mistral 7           & 28\%  & 13 min \\
  Ollama - Phi 3.5             & 22\%  & 9 min  \\
  Ollama - LLaMA 3.2           & 6\%   & 9 min  \\
  \hline
  \end{tabular}
  \caption{Preliminary evaluation of LLMs on the posology structuration task. The latency is measured across the whole test dataset. Due to the closed-source nature of Gemini models Ollama's and the NERL's latency is not directly comparable with Vertex AI, as it does not run on the same hardware. However, the latency still provides a useful indication of what is possible with readily available models and hardware.}
  \label{tab:llm-eval}
\end{table}

From these initial results, Gemini 1.5 Flash emerges as the most promising
candidate, offering a good balance between performance and latency. While Gemini
1.5 Pro and Gemma 2 are slightly outperforming in terms of accuracy, their
significantly higher inference time makes them impractical for iterative
experimentation. 

Once the model is selected, we proceed with iterative prompt refinement. Importantly, we observe that prompt engineering is highly model-specific; optimized prompts used for one LLM do not transfer effectively to others. Therefore, we proceed with development and evaluation of prompt strategies exclusively for Gemini 1.5 Flash~\footnote{Specifically, we evaluated the final prompt optimized for Gemini 1.5 Flash on other models. Gemini 1.5 Pro, despite outperforming Flash on the initial prompt, achieved 8\% lower accuracy than Flash when using the Flash-optimized version.}. This is consistent with previous literature, which shows that engineered prompts often do not transfer well across different models \cite{ye-etal-2024-prompt}.

We begin with relatively simple prompting techniques~\footnote{Further details on prompt design are provided in Appendix A}, including direct task instructions, chain-of-thought reasoning, and few-shot examples embedded in the prompt. While these approaches improve performance, they do not fully resolve the inherent ambiguities of medical language. To mitigate this, we introduce rephrase-and-respond instructions, guiding the model to first reformulate the input to clarify intent before attempting to structure it. Finally, we apply contrastive prompting, where examples of incorrect outputs and common mistakes were explicitly included to guide the model away from frequent failure patterns.  

All prompts are executed with the temperature set to zero, ensuring deterministic and consistent output generation. We empirically observed that setting the temperature to zero led to improved accuracy, as shown in Table~\ref{tab:params}. This is consistent with previous findings that lower temperatures have a tendency to produce more repetitive and less diverse outputs, which resemble the training data more closely \cite{renze-2024-effect}.

\begin{table}[ht]
  \centering
  \begin{tabular}{rrrr}
  \hline
\textbf{top k} & \textbf{top p} & \textbf{temperature} & \textbf{Average Accuracy} \\
  \hline
  40  & 0.80   & 2.0   & 68.75   \\
  35  & 0.90   & 1.5   & 67.19   \\
  30  & 0.85   & 1.2   & 68.75   \\
  15  & 0.90   & 1.0   & 68.75   \\
  5   & 0.97   & 0.7   & 71.88   \\
  5   & 1.00   & 0.1   & 73.00   \\
  -   & -  & 0.0   & 73.00   \\
  \hline
  \end{tabular}
  \caption{Evaluation of different parameter settings for the Gemini 1.5 Flash
  model. The parameters are: \textbf{top k} (number of top tokens to consider),
  \textbf{top p} (nucleus sampling threshold), and \textbf{temperature}
  (controls randomness in output generation).}
  \label{tab:params}
\end{table}

Using this fully engineered prompt set, Gemini 1.5 Flash achieves an accuracy of 73\%. However, despite the gains, prompt engineering alone still falls short of our NERL baseline. The results motivate the application of fine-tuning.

We fine-tune the model with PEFT method available throuth Vertex AI using a filtered synthetic training dataset of 1~200 high-confidence outputs generated by our NERL pipeline. The tuning is performed in two stages. First, the model is trained on plain input-output pairs, resulting in an accuracy of 79\%. In the second stage, we include a partial version of the final prompt within the training inputs: general task instructions, a chain-of-thought example, and a list of mistakes to avoid. This alignment leads to a further improvement, reaching 84\% accuracy and approaching our NERL baseline. Final results are shown in Table~\ref{tab:llm-results}.

\begin{table}[h]
  \centering
  \begin{tabular}{lrr}
  \hline
  \textbf{Model Configuration} & \textbf{Average Accuracy} & \textbf{Average Latency} \\
  \hline
  Gemini 1.5 Flash (prompt-engineered) & 73\% & 3 min \\
  Gemini 1.5 Flash (fine-tuned)        & 79\% & 3 min \\
  Gemini 1.5 Flash (prompt-engineered + fine-tuned)        & 84\% & 3 min \\
  NERL baseline                        & 85\% & 1 min \\
  \hline
  \end{tabular}
  \caption{Final evaluation of LLMs on the posology structuration task. Please note that the average latency for the NERL baseline is not directly comparable to the LLMs, as it does not run on the same hardware.}
  \label{tab:llm-results}
  \end{table}

  In summary, our experiments demonstrate that while prompt engineering can significantly improve LLM performance on complex, ambiguous tasks like posology structuration, only model-specific fine-tuning enables LLMs to reach the performance levels of a robust baseline.

\subsection{Error analysis}

To compare the performance of NERL and Gemini 1.5 Flash we conduct an error analysis. Both models demonstrate high performance in structuring posologies, but diverge in their handling of linguistic complexity and ambiguity.

NERL, for instance, struggles with implicit or colloquial conditional expressions. For example, it fails to extract \textit{si besoin} from \textit{Lactulose 10 g: si besoin}, and similarly misses \textit{si douleurs} as a conditional intake trigger. It also shows weakness in detecting and normalizing OCR errors, for example, \textit{sachet} in \textit{1 sach dose...}. Additionally, unit recognition is sometimes inconsistent, particularly when syntaxis is atypical, as seen in \textit{Comprimé : 2 à 20:00}, where \textit{comprimé} was not extracted.

In contrast, Gemini tends to add unnecessary abstractions, like extracting \textit{period: 1} and \textit{period unit: day} from \textit{prendre 1 comprimé 3 fois par 24 heures} or refomatting, like transforming \textit{15/11/2022} into \textit{2022/11/15}. It also has difficulty parsing multi-timepoint instruction: in \textit{2 gélules au cours ou immédiatement après les repas}, it confuses overlapping timing cues (like \textit{during the meal} vs. \textit{after a meal}) and returns only one. Gemini also misinterpretes dose allocation across multiple intake times: \textit{Prendre 2 gélules le matin et le soir...} was interpreted as 2 capsules once daily, instead of once in the morning and once in the evening.

In conclusion, while NERL and Gemini 1.5 Flash each show specific weaknesses, they also demonstrate complementary strengths. NERL favors precision and structural consistency, whereas Gemini excels in semantic interpretation and normalization. Used together, these models could provide a more robust and comprehensive solution for structuring posology task.

\subsection{Towards an LLM-based hybrid system}

Rather than replacing NERL, we propose a hybrid approach based on confidence score. The workflow we propose begins with processing the input document using NERL. If NERL’s confidence score falls below a defined threshold (<0.8), the task is delegated to the LLM for supplementary analysis. The final output is selected based on the highest confidence score between the two systems.

This hybrid strategy leverages the complementary strengths of NERL and the LLM. Moreover, the approach minimizes latency and computational overhead by limiting LLM invocation to low-confidence cases. Results are shown in Table~\ref{tab:hybrid-llm-results}, and demonstrate that the hybrid system outperforms both NERL and LLM alone, achieving an average accuracy of 91\%.

\begin{table}[h]
  \centering
  \begin{tabular}{lr}
  \hline
  \textbf{Model Configuration} & \textbf{Average Accuracy}\\
  \hline
  Gemini 1.5 Flash (fine-tuned)        & 84\% \\
  NERL baseline                        & 85\% \\
  Hybrid system (LLM + NERL)           & 91\% \\
  \hline
  \end{tabular}
  \caption{Final evaluation of the hybrid system on the posology structuration task.}
  \label{tab:hybrid-llm-results}
\end{table}

\vspace{-0.5cm}

\section{Conclusion}

This work introduces \dataset, a novel dataset for evaluating the LLMs ability
to convert raw textual French posology instructions into a structured format.
This task is more challenging than standard NER, as it requires to generate a
full JSON output rather than simply tagging spans of text.

To address the limitations of both our NERL baseline and the LLMs on this task,
we proposed a hybrid strategy that leverages a surrogate confidence score to
combine their outputs, offering a practical and efficient solution for
real-world deployment.

Our results also suggest that proprietary LLMs hosted on cloud platforms allow
prompt engineering and fine-tuning with less computational resources and less
engineering and they perform better than open-source models at comparable levels
of latency. However, with more resources than for this work, future work could
explore the possibility of using smaller LLMs with domain-specific data, like
BioMistral \cite{labrak-etal-2024-biomistral}, and with guided generation
\cite{willard2023efficientguidedgenerationlarge} to improve the performance of open-source models.

%%================================================================
%% Note : si l'on préfère éviter de factoriser les crossrefs :
%% bibtex -min-crossrefs 99 taln-exemple
%%================================================================
\bibliographystyle{coria-taln2025}
\bibliography{biblio}

\clearpage

\appendix

\section{Acknowledgements}

We would like to thank Eliott Tourtois for his help in the annotation of the
dataset, and in the conception of the structuration schema. We would also like
to thank all the colleagues who contributed to the development of the NERL
baseline, in particular, Patricio Cerda, Xavier Fontaine, and Baptiste Charnier.
We are also grateful for the entire team of Posos for their support and for
their indirect contribution to this work. Finally, we would like to thank the
reviewers for their valuable comments and suggestions.

\section{Complete structuration schema}\label{sec:schema}

\begin{itemize}
\item \texttt{as\_needed}: dict (optional), contains information about whether the drug must be taken only under certain condition (e.g. if "as needed during headache" is written)
\begin{itemize}
  \item \texttt{as\_needed}: bool, True if the drug must be taken under certain condition
  \item \texttt{as\_needed\_for}: str, the specific condition (e.g. "during headache", "if needed", "if pain")
\end{itemize}
\item \texttt{designation}: str, the minimal string to which the dosage corresponds, it does not contain the whole instruction, but rather the unit and the type of intake (e.g. "1 comprimé"), this allows to differentiate several posology instructions when needed
\item \texttt{quantity\_and\_rate}: dict, contains the amount of intake units to take at once
\begin{itemize}
  \item \texttt{value}: float, the amount of unit to take at once
  \item \texttt{unit}: str (optional), the type of unit  (e.g. "comprimé(s)", "ampoule(s)", "sachet(s)", "cmp", "cp", "sach", "amp", etc.)
  \item \texttt{code}: str (optional), the SNOMED code for the unit
\end{itemize}
\item \texttt{max\_dose\_per\_period}: dict (optional), contains information about a potential maximum dosing
\begin{itemize}
  \item \texttt{dose}: int, the maximum amount of units
  \item \texttt{dose\_unit}: str (optional), the type of unit
  \item \texttt{code}: str (optional), the SNOMED code of the dose\_unit
\end{itemize}
\item \texttt{timing}: dict, contains the timing with which each intake must take place
\begin{itemize}
  \item \texttt{bounds\_duration}: dict (optional), the duration during which the treatment occurs
  \begin{itemize}
    \item \texttt{max\_value}: float (optional), the maximum amount of time units the treatment must last
    \item \texttt{value}: float, the amount of time units the treatment must last (if max\_value is not null, value reperesent the minimal value)
    \item \texttt{unit}: str, the unit for measuring the treatment duration ("hours", "day", "week", or "month")
  \end{itemize}
  \item \texttt{bounds\_period}: dict (optional), an alternative to bounds\_duration where exact dates have been provided
  \begin{itemize}
    \item \texttt{start\_date}: str, start date in format YYYY/MM/DD
    \item \texttt{end\_date}: str, end date in format YYYY/MM/DD
  \end{itemize}
  \item \texttt{day\_of\_week}: list (optional), list of weekday diminutives when the treatment must be taken (e.g. "mon" or "sat")
  \item \texttt{frequency}: int, an integer for the number of intake for the given period (e.g. the 2 in "twice a day")
  \item \texttt{frequency\_max}: int (optional), an integer for the maximum amount fo intake for the given period (frequency is then the minimum)
  \item \texttt{number\_repeats\_allows}: int (optional), the number of times the treatment can be renewed
  \item \texttt{offset}: str (optional), substring extracted from the query that indicates what is the offset of the intake ("30 minutes" in "30 minutes before meals")
  \item \texttt{period}: int, the number of period\_unit to observe before repeating the intake (6 in "3 tablets every 6 hours")
  \item \texttt{period\_unit}: str, the time unit used for measuring the period between intakes ("hours" in "3 tablets every 6 hours")
  \item \texttt{sequence}: int (optional), when several posology instructions are given, sequence indicates their order
  \item \texttt{time\_of\_day}: list (optional), indicates precise intake hours in the format HH:mm:ss. \texttt{time\_of\_day} and \texttt{when} cannot be simultanously non-empty
  \item \texttt{when}: list (optional), indicates the moment of intake relative to the patient activity (e.g. "AC" for before a meal), the list of possible values is given in the FHIR standard\footnote{\url{https://hl7.org/fhir/R4/valueset-event-timing.html}}
\end{itemize}
\end{itemize}

\section{Dataset statistics}

\begin{table}
  [ht]
  \centering
  \begin{tabular}{lr}
  \hline
  \multicolumn{2}{l}{\textbf{Statistics}} \\
  \hline
  Number of queries   & 129 \\
  Number of posology instructions  & 131 \\
  Minimum number of tokens & 2 \\
  Median number of tokens & 9 \\
  Maximum number of tokens & 29 \\
  Minimum number of occurrences of a JSON field & 1 \\
  Median number of occurrences of a JSON field & 69\\
  Maximum number of occurrences of a JSON field & 131 \\
  \hline
  \end{tabular}
  \caption{Statistics of the \dataset dataset.}
  \label{tab:data_statistics}
\end{table}

\section{NER training dataset}\label{sec:ner-training}

\begin{table}
  [ht]
  \centering
  \begin{tabular}{lrl}
  \hline
  \textbf{Entity types} & \textbf{Number} & \textbf{Example} \\
  \hline
  \texttt{BOUNDS} & 647 & pendant 7 jours \\
  \texttt{DOSE} & 1,188 & 2cp \\
  \texttt{DRUG} & 1,007 & EFFICORT Hydrophile \\
  \texttt{FORM} & 574 & CPR \\
  \texttt{FREQUENCY} & 213 & 3 fois \\
  \texttt{NUMBER\_REPEATS} & 20 & 1 Boîte.a renouveler \\
  \texttt{PERIOD} & 555 & par jour \\
  \texttt{REASON} & 143 & SI BESOIN \\
  \texttt{STRENGTH} & 721 & 200 mg \\
  \texttt{TIME\_OF\_DAY} & 19 & à 08h \\
  \texttt{WHEN} & 723 & le matin \\
  \hline
  \end{tabular}
  \caption{NER entities present in the training dataset}
  \label{tab:ner_entities}
\end{table}

\section{Prompt details}\label{sec:prompt-details}

\paragraph{1. Chain of thought examples}\mbox{}\

here is an exmaple 1.1 how to proceed.

sentence: \texttt{ampoute , 1 fois par trimestre besoin}

step by step reasonning: 

0. Check spelling, remove punctuation and make the sentence more explicit: \texttt{ampoute , 1 fois par trimestre besoin} most likely means \texttt{1 ampoule 1 fois par trimestre si besoin}

1. Check whether there are several entities. Here, ampoute (misspelled) refers to only one entity.

2. I see that the sentence refers to medicines, so I can associate \texttt{category} with \texttt{MEDICATION}.

3. Since there is only one entity, the \texttt{designation} can be associated with the value of the entire sentence: \texttt{ampoute 1 fois par trimestre besoin}

4. We can see that the sentence talks about ampoute (misspelled), 1 time per trimester, so it's an \texttt{entity\_type} to be associated with the \texttt{QUANTITY} value. 

5. The \texttt{quantity\_and\_rate} will contain the values \texttt{type} associated with \texttt{DOSE} (one ampoule corresponds to one dose of product), {unit} associated with \texttt{ampoule(s)} and \texttt{value} associated with \texttt{1} (because once per quarter in the sentence). There are no elements relating to the maximum value, so there's no need to set \texttt{max\_value}.

6. The \texttt{timing} section. The value of \texttt{bounds\_duration\_text} will be equal to \texttt{1 fois par trimestre}, which means that the \texttt{frequency} will be equal to \texttt{1}, the \texttt{period} to \texttt{3} and the \texttt{period\_unit} to \texttt{month} because a quarter is equal to 3 months.

7. Concerning the \texttt{as\_needed} section, as it writes \texttt{besoin} which probably means \texttt{si besoin}, we'll set the \texttt{as\_needed} value to \texttt{True} and the \texttt{as\_needed\_for} value to \texttt{besoin}.

The result is the following YAML format:

\begin{verbatim}
  entities:
  - a_needed:
      as_needed: true
      as_needed_for: besoin
    category: MEDICATION
    designation: ampoute 1 fois par trimestre besoin
    entity_type: QUANTITY
    quantity_and_rate:
      type: DOSE
      unit: ampoule(s)
      value: 1
    timing:
      bounds_duration_text: ''
      frequency: 1
      frequency_texts:
      - 1 fois par trimestre
      period: 3
      period_unit: month
  posology_string: ampoute 1 fois par trimestre besoin
\end{verbatim}

\paragraph{2. Few-shot examples}\mbox{}\

here is an exmaple 2.1 how to proceed.

sentence: \texttt{0.5 à 1 cp au coucher si besoin (1 boite AR)}

\begin{verbatim}
  entities:
  - category: MEDICATION
    designation: 0.5 a 1 cp au coucher si besoin
    as_needed:
      as_needed: true
      as_needed_for: si besoin
    entity_type: QUANTITY
    quantity_and_rate:
      max_value: 1.0
      type: DOSE
      unit: comprimé(s)
      value: 0.5
    timing:
      bounds_duration_text: ''
      frequency: 1
      frequency_texts:
      - au coucher
      period: 1
      period_unit: day
      when:
      - HS
  posology_string: 0.5 à 1 cp au coucher si besoin
\end{verbatim}

\paragraph{3. Mistakes to avoid}\mbox{}\

If a time of day is specified in hours, don't forget to specify it in the \texttt{time\_of\_day} field (even if the when field is also filled)

Be careful with time of intake: \texttt{le matin 30 minutes avant le repas} is not two different time of day for in the morning and before meal (\texttt{MORN}, \texttt{AC}) but it represents a single one for taking the medication before the meal of the morning (\texttt{ACM})

For time of days that fit in the \texttt{when} field, do not mistake meals (like \texttt{ACM} for breakfast and \texttt{CD} for lunch) with periods of the day (like \texttt{MORN} for morning and \texttt{NOON} for mid-day), you must use a meal time only if the meal is explicitly described

Be careful with \texttt{as\_needed} field, it should include \texttt{as\_needed} equals \texttt{True} and \texttt{as\_needed\_for} referencing condition, if you detect relevant condition in text for the entity in question (\texttt{si besoin}, \texttt{si migraine}, \texttt{si angoisse}). \texttt{as\_needed} field should not constitute a separate entity.

%%================================================================
\end{document}